\documentclass[11pt,a4paper]{article}
\usepackage[hyperref]{rocling2025}
\usepackage{times}
\usepackage{latexsym}
\usepackage{multirow}
\usepackage{booktabs}
\usepackage{amsmath}
\usepackage{hyperref}

\usepackage{fontspec}
\usepackage{microtype}
\usepackage{graphicx}
\usepackage{tabularx}

\roclingfinalcopy

\title{CLiFT-ASR: A Cross-Lingual Fine-Tuning Framework for Low-Resource Taiwanese Hokkien Speech Recognition}

\author{
\begin{tabular}{c}
Hung-Yang Sung\textsuperscript{1}, Chien-Chun Wang\textsuperscript{1}, 
Kuan-Tang Huang\textsuperscript{1}, Tien-Hong Lo\textsuperscript{1}, \\
Yu-Sheng Tsao\textsuperscript{2}, Yung-Chang Hsu\textsuperscript{2}, 
and Berlin Chen\textsuperscript{1}
\end{tabular} \\
\textsuperscript{1}National Taiwan Normal University, Taiwan \\
\textsuperscript{2}EZAI, Taiwan \\
\texttt{\{redsheep, 61247033s, 61347002s, teinhonglo, berlin\}@ntnu.edu.tw}, \\
\texttt{\{sam, mic\}@ez-ai.com.tw}
}

\date{}

\begin{document}
\maketitle

\begin{abstract}

Automatic speech recognition (ASR) for low-resource languages such as Taiwanese Hokkien is difficult due to the scarcity of annotated data.
However, direct fine-tuning on Han-character transcriptions often fails to capture detailed phonetic and tonal cues, while training only on romanization lacks lexical and syntactic coverage.
In addition, prior studies have rarely explored staged strategies that integrate both annotation types.
To address this gap, we present CLiFT-ASR, a cross-lingual fine-tuning framework that builds on Mandarin HuBERT models and progressively adapts them to Taiwanese Hokkien.
The framework employs a two-stage process in which it first learns acoustic and tonal representations from phonetic Tai-lo annotations and then captures vocabulary and syntax from Han-character transcriptions.
This progressive adaptation enables effective alignment between speech sounds and orthographic structures.
Experiments on the TAT-MOE corpus demonstrate that CLiFT-ASR achieves a 24.88\% relative reduction in character error rate (CER) compared with strong baselines.
The results indicate that CLiFT-ASR provides an effective and parameter-efficient solution for Taiwanese Hokkien ASR and that it has potential to benefit other low-resource language scenarios.

\end{abstract}


\section{Introduction}

Taiwanese Hokkien is an important dialect in Taiwan with rich cultural and historical significance.
However, as Mandarin Chinese dominates education and daily life, the use of Taiwanese Hokkien has been declining, especially among younger generations.
A 2020 survey\footnote{\url{https://www.stat.gov.tw/News_Content.aspx?Create=1&n=2755&state=1327FD6AD8DCDA52&s=230300&ccms_cs=1&sms=11065/}} reports that only 7.4\% of children regularly use Taiwanese Hokkien.
Despite the existence of several speech corpora \cite{liao2022,chou2023, lin2024}, the overall amount of annotated data is limited compared to high-resource languages such as Mandarin and English \cite{zhang2022,wang2021}.
This data scarcity poses a significant challenge for developing robust Speech Translation \cite{chen2023} and automatic speech recognition (ASR) systems.

Existing Taiwanese Hokkien ASR systems face additional challenges due to inconsistent transcription standards.
Some systems employ Tai-lo romanization \cite{chou2023,chao2021}, which combines phonetic scripts with tonal markings, making it less intuitive and harder for general users to accept \cite{khoo2019}.
Other approaches annotate speech with Mandarin characters, but the mapping between Taiwanese Hokkien vocabulary and Mandarin text is often one-to-many or partially aligned, leading to longer and less accurate output sequences \cite{lin2024}.
Using Taiwanese Hokkien Han characters provides a practical alternative that balances readability and phonological detail, improving recognition usability.

To overcome these challenges, we introduce \textbf{CLiFT-ASR}\footnote{\url{https://github.com/redsheep913/CLiFT-ASR/}}, a \textbf{C}ross-\textbf{Li}ngual \textbf{F}ine-\textbf{T}uning framework for low-resource \textbf{A}utomatic \textbf{S}peech \textbf{R}ecognition that leverages Mandarin HuBERT backbone models and progressively adapts them to Taiwanese Hokkien.
The framework follows a two-stage fine-tuning strategy where it first acquires acoustic-level knowledge from phonetic Tai-lo annotations and then learns language-level structures such as vocabulary and syntax using Taiwanese Hokkien Han characters.
Comprehensive experiments on the TAT-MOE corpus demonstrate that CLiFT-ASR achieves a 24.88\% relative reduction in character error rate (CER).
The framework offers an effective solution for Taiwanese Hokkien ASR and provides guidance for developing ASR systems for other low-resource languages.

\section{Background}

\subsection{Linguistic Characteristics of Taiwanese Hokkien}

Taiwanese Hokkien has a seven-tone system and exhibits tone sandhi, which creates tonal variations that differ from Mandarin \cite{cheng1968}.
These tonal patterns make automatic speech recognition challenging, as accurate recognition requires modeling both static tones and context-dependent tone changes.
Despite these differences, Taiwanese and Mandarin share similar morphological and syntactic structures \cite{sun2006}, which allows knowledge transfer from Mandarin-pretrained models.
Previous studies show that Mandarin-pretrained ASR models outperform English-pretrained models when recognizing romanized Taiwanese (Tai-lo), indicating that cross-lingual transfer can be effective for end-to-end ASR targeting Taiwanese Han characters \cite{chou2023}.
These observations motivate the design of CLiFT-ASR, which leverages cross-lingual knowledge and adapts it progressively to Taiwanese Hokkien.
Note that this work does not focus on modeling tone sandhi phenomena, which is left for future research.

\subsection{Orthographic Systems and Their Role in ASR}

Taiwanese Hokkien uses two main orthographies: romanization and Han characters.
Romanization systems such as Pe̍h-ōe-jī (POJ) and Tai-lo provide systematic phonetic representations \cite{khoo2019}.
The Ministry of Education has published a recommended set of roughly 700 Han characters for writing Taiwanese, which can be combined with romanization in a mixed-script form known as hàn-lô \footnote{\href{https://language.moe.gov.tw/001/Upload/files/site_content/M0001/language_100/D/D005.pdf}{\nolinkurl{https://language.moe.gov.tw/.../D005.pdf}}}.
For ASR, using Han characters or hàn-lô offers a practical balance between phonetic detail and readability and informs the two-stage fine-tuning strategy employed in CLiFT-ASR.

\begin{figure}[t]
\centering
\includegraphics[width=0.95\linewidth]{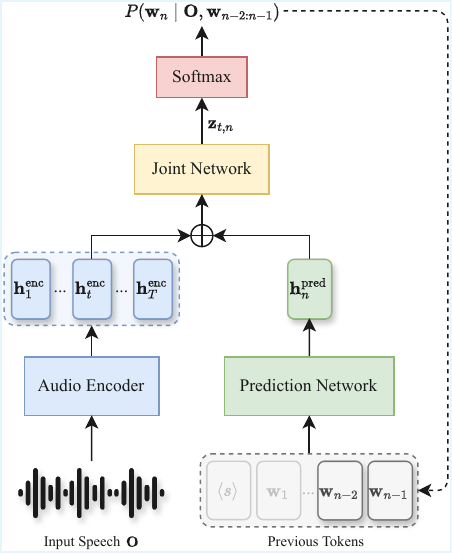}
\caption{
Overview of the proposed CLiFT-ASR.
The $\bigoplus$ operator denotes element-wise tensor addition.
The dashed arrow indicates that during inference, ground-truth labels are not available, so the model outputs are fed back autoregressively into the prediction network.
}
\label{fig:rnnt}
\vspace{-10pt}
\end{figure}

\section{Proposed Method}

\subsection{Model Architecture}

The proposed CLiFT-ASR framework builds upon the RNN-Transducer (RNN-T) \cite{graves2012} to align variable-length acoustic sequences with token sequences, as illustrated in Figure~\ref{fig:rnnt}.
Given an input audio signal $\mathbf{O}$ and a sequence of target tokens $\mathbf{W} = (\mathbf{w}_1, \mathbf{w}_2, \dots, \mathbf{w}_N)$, where $N$ denotes the number of output tokens, CLiFT-ASR estimates a probability distribution over possible tokens at each alignment step.
The model consists of three components: an audio encoder, a prediction network, and a joint network.
The audio encoder processes $T$ acoustic feature frame vectors $(\mathbf{o}_1, \mathbf{o}_2, \dots, \mathbf{o}_T)$ extracted from $\mathbf{O}$ and maps them to a sequence of high-level representations that capture phonetic, tonal, and other essential speech information:
\begin{equation}
\mathbf{H}^{\text{enc}} = \operatorname{AudioEncoder}(\mathbf{O}),
\end{equation}
where $\mathbf{H}^{\text{enc}}$ represents the sequence of encoder hidden states $(\mathbf{h}_1^{\text{enc}}, \dots, \mathbf{h}_T^{\text{enc}})$.
This representation integrates both local and global acoustic patterns, which is crucial for accurately modeling the tonal variations in Taiwanese Hokkien.
The prediction network generates a context representation autoregressively, conditioning on the previous two target tokens to form a trigram-style context that models short-term sequential dependencies in the output space:
\begin{equation}
\mathbf{h}_n^{\text{pred}} = \operatorname{PredictionNetwork}(\mathbf{w}_{n-2}, \mathbf{w}_{n-1}).
\end{equation}
The joint network combines the encoder output at time step $t$, 
$\mathbf{h}_t^{\text{enc}}$, which corresponds to the $t$-th element of $\mathbf{H}^{\text{enc}}$, 
with the prediction network state to form a joint representation. 
The conditional distribution over the next token is then obtained by applying a softmax:
\begin{equation}
\mathbf{z}_{t,n} = \operatorname{JointNetwork}(\mathbf{h}_t^{\text{enc}} + \mathbf{h}_n^{\text{pred}}),
\end{equation}
\begin{equation}
P(\mathbf{w}_n \mid \mathbf{O}, \mathbf{w}_{n-2:n-1}) = \operatorname{Softmax}(\mathbf{z}_{t,n}).
\end{equation}
This architecture allows CLiFT-ASR to jointly leverage acoustic and linguistic context at each step, which is essential for capturing tonal and phonological patterns in Taiwanese Hokkien.

\subsection{Training Strategy}

To handle limited annotated Taiwanese Hokkien data, CLiFT-ASR adopts a two-stage fine-tuning framework based on a pre-trained Mandarin HuBERT encoder.
In the first stage, the model learns acoustic-level representations from phonetic Tai-lo annotations.
Given a training set $\{(\mathbf{O}^{(i)}, \mathbf{W}^{(i)}_{\text{Tai-lo}})\}_{i=1}^{U_{\text{Tai-lo}}}$, the model parameters $\theta$ are updated to minimize the negative log-likelihood:
\begin{equation}
\theta' = \arg\min_{\theta} \sum_{i=1}^{U_{\text{Tai-lo}}} - \log P\big(\mathbf{W}^{(i)}_{\text{Tai-lo}} \mid \mathbf{O}^{(i)}; \theta\big).
\end{equation}
This stage captures fine-grained acoustic and phonetic details, providing a solid foundation for language-level learning.
In the second stage, the model is fine-tuned on Taiwanese Hokkien Han character annotations. Starting from the network parameter $\theta'$, it is trained on $\{(\mathbf{O}^{(j)}, \mathbf{W}^{(j)}_{\text{Han}})\}_{j=1}^{U_{\text{Han}}}$ to learn vocabulary, syntax, and higher-level linguistic structures:
\begin{equation}
\theta^* = \arg\min_{\theta} \sum_{j=1}^{U_{\text{Han}}} - \log P\big(\mathbf{W}^{(j)}_{\text{Han}} \mid \mathbf{O}^{(j)}; \theta\big).
\end{equation}
By progressively adapting from phonetic to linguistic representations, CLiFT-ASR effectively leverages cross-lingual knowledge and maximizes the use of limited annotated data, resulting in more accurate and robust Taiwanese Hokkien ASR.

\begin{table}[t]
\centering
\setlength{\tabcolsep}{7pt}
\begin{tabular}{lcccc}
\toprule
\textbf{Split} & \textbf{Spk.} & \textbf{Utt.} & \textbf{Hr.} \\
\toprule
Training & 328 & 86,072 & 153.33 \\
Development & 58 & 16,357 & 28.60  \\
Test & 54 & 15,962 & 26.28  \\
\midrule
Total & 440 & 118,391 & 208.21 \\
\bottomrule
\end{tabular}
\caption{Statistics of the TAT-MOE dataset across training, development, and test splits, including the number of speakers (Spk.), utterances (Utt.), and total duration in hours (Hr.).}
\label{tab:tatmoe}
\vspace{-10pt}
\end{table}

\begin{table*}[t]
\centering
\setlength{\tabcolsep}{6pt}
\begin{tabular}{lccccccc}
\toprule
\multirow{2}{*}{\textbf{Model}} & \multirow{2}{*}{\textbf{Parameters (M)}} & \multicolumn{2}{c}{\textbf{Development}} & \multicolumn{2}{c}{\textbf{Test}} & \multicolumn{2}{c}{\textbf{Clean Test}} \\
\cmidrule(lr){3-4}
\cmidrule(lr){5-6}
\cmidrule(lr){7-8}
 & & \bf{CER} & \bf{Rel.} & \bf{CER} & \bf{Rel.} & \bf{CER} & \bf{Rel.} \\
\toprule
Zipformer & 65 & 48.57 & - & 45.82 & - & 15.69 & - \\
FSR-2020 Best & - & - & - & - & - & 15.62 & 0.07 \\
Whisper-base & 74 & 27.36 & 21.21 & 24.02 & 21.80 & 10.05 & 5.64 \\
HuBERT-base & 96 & 26.16 & 22.41 & 24.49 & 21.33 & 12.97 & 2.72  \\
HuBERT-base-cmn & 96 & 24.06 & 24.51 & 22.41 & 23.41 & 9.08 & 6.61 \\
\bf{CLiFT-ASR} & \bf{96} & \bf{22.37} & \bf{26.20} & \bf{20.94} & \bf{24.88} & \bf{8.06} & \bf{7.63} \\
\midrule
Whisper-small & 244 & 22.47 & 26.10 & 18.68 & 27.14 & 7.66 & 8.03 \\
\bottomrule
\end{tabular}
\caption{
CERs (\%) and relative reductions (Rel., \%) for Taiwanese Hokkien ASR using various audio encoder initialization strategies.CLiFT-ASR applies a two-stage fine-tuning strategy with the HuBERT-base-cmn encoder.
FSR-2020 Best refers to the top-performing model from FSR-2020.
}
\label{tab:init}
\vspace{-10pt}
\end{table*}

\section{Experimental Setup}

\subsection{Dataset}

All experiments were conducted on the TAT-MOE subset of the TAT corpus \cite{liao2022}, a large-scale Taiwanese Hokkien speech resource covering diverse regions of Taiwan.
The corpus captures variation in speaker accents and pronunciation, providing a suitable testbed for robust ASR development.
Audio recordings were sampled at 16 kHz with 16-bit PCM encoding to ensure consistent acoustic quality.
Transcriptions were provided in Hàn-Lô-Tâi-bûn, a mixed orthography combining Han characters and romanized phonetics.
Alternative annotations, including Pe̍h-ōe-jī, Tai-lo, and tone-marked Tai-lo, were also available to support different modeling strategies.
Table~\ref{tab:tatmoe} summarizes the number of speakers, utterances, and total duration for the training, development, and test sets.
To further evaluate model performance, we included a cleaner test set drawn from the pilot test of the Formosa Speech Recognition Challenge 2020 (FSR-2020) \cite{liao2020}, referred to as the clean test.
The TAT-MOE corpus therefore provides high-quality acoustic data and multiple orthographic representations, making it a valuable benchmark for low-resource Taiwanese Hokkien ASR.

\subsection{Data preprocessing}

The transcripts in the TAT-MOE dataset were written in Hàn-Lô-Tâi-bûn, a mixed system of Han characters and romanized phonetics. 
To unify the representation, we first constructed a mapping table using additional corpora to convert romanized segments into the corresponding Han characters. 
Arabic numerals were also converted into Chinese numerals, and variant or synonymous characters were normalized to a single standardized form. 
These preprocessing steps reduce inconsistencies and lexical variation in the annotations, thereby improving the stability of training and the accuracy of recognition.

\subsection{Model Configuration}

All training procedures followed Icefall's official recipes and default settings\footnote{\url{https://github.com/k2-fsa/icefall/}}. 
To establish a fair baseline and assess the benefit of cross-lingual transfer, we considered two encoder configurations: the baseline Zipformer model and a HuBERT-based Transformer initialized with Mandarin pretrained weights provided by the toolkit. 
The prediction network adopted Icefall's stateless design for efficient sequence modeling, and the joint network followed the standard implementation for integrating audio encoder and prediction network features into output distributions \cite{yao2024,hsu2021,ghodsi2020}. 
For tokenization, we employed Icefall's byte-level BPE model, which has proven effective for handling large CJK vocabularies in bilingual and multilingual ASR tasks. %之前有cite也是byte-level BPE 
This configuration enables a direct comparison between a strong baseline and our cross-lingual strategy, ensuring that performance gains are consistent and interpretable.

\subsection{Training Details}

Speech data were prepared using the Lhotse toolkit \cite{zelasko2021}.
For feature extraction, the Zipformer baseline model used 80-dimensional filter bank (FBank) features, while the HuBERT-based model was fine-tuned directly from raw waveform inputs.
In CLiFT-ASR, the first stage was trained for 20 epochs and the second stage for 40 epochs. For comparison, a direct fine-tuning approach without staging was trained for 60 epochs. All models were trained with gradient accumulation over 4 steps to stabilize optimization. 
To balance computational efficiency and contextual coverage, the maximum audio duration per training sample was limited to 120 seconds. 
The learning rate was initialized at 0.0005 and scheduled over 40 epochs for smooth convergence. 
Model embeddings were set to 256 dimensions, and training was initialized from pretrained checkpoints. 
Optimization was performed with the ScaledAdam optimizer, which applied adaptive learning rates and gradient clipping at 2.0 for stability. 
A custom learning rate scheduler, Eden, was employed to dynamically adjust the learning rate across both batch and epoch progression \cite{yao2024}.

\begin{table*}[t]
\centering
\setlength{\tabcolsep}{7.5pt}
\begin{tabular}{lcccc}
\toprule
\textbf{Fine-tuning Strategy} & \textbf{Frozen} & \textbf{Development} & \textbf{Test} & \textbf{Clean Test} \\
\toprule
Direct & None & 24.06 & 22.41 & 9.08  \\
\midrule
\multirow{4}{*}{\bf{Two-stage}}  
& Audio Encoder & 36.82 & 35.72 & 26.84 \\
& Prediction Network & 25.23 & 23.91 & 11.84 \\
& Joint Network & 29.58 & 28.59 & 18.46 \\
& \bf{None} & \bf{22.37} & \bf{20.94} & \bf{8.60} \\
\bottomrule
\end{tabular}
\caption{
CERs (\%) on development, test, and clean test sets for different training strategies and parameter freezing configurations.
The table compares direct fine-tuning with the proposed two-stage strategy, evaluating the impact of freezing specific components (audio encoder, prediction network, joint network) during the first stage.
}
\label{tab:finetune}
\vspace{-10pt}
\end{table*}

\subsection{Evaluation Metric}

Character error rate (CER) was employed as the primary evaluation metric. 
CER quantifies the discrepancy between the predicted output and the reference transcription by counting the number of substitutions, deletions, and insertions. 
It is computed as the ratio of total character errors to the number of characters in the reference:
\begin{equation}
\text{CER} = \frac{S + D + I}{C},
\end{equation}
where $S$, $D$, and $I$ represent the numbers of substitutions, deletions, and insertions, respectively, and $C$ denotes the total number of characters in the reference.
As a character-level measure, CER provides a precise and widely accepted evaluation of recognition accuracy for speech recognition tasks, with lower values indicating better performance. 
For Taiwanese Hokkien, where annotations include a mix of Han characters and romanized phonetics, CER is particularly suitable because it captures errors across both orthographic forms and effectively reflects the ability of the model to handle tonal and phonological variations.

\section{Results and Discussion}

\subsection{Effects of Language Initialization}

Table~\ref{tab:init} summarizes the impact of different encoder initialization strategies on Taiwanese Hokkien ASR performance.
The comparison includes Zipformer without pretraining, HuBERT-base pretrained on English, Whisper models with multilingual pretraining, and HuBERT-base-cmn pretrained on Mandarin.
CLiFT-ASR, built on the Mandarin-pretrained HuBERT-base-cmn encoder and the proposed two-stage fine-tuning strategy, achieves the strongest overall performance.

Models without language-specific pretraining, such as Zipformer, exhibit the lowest performance, highlighting the difficulty of learning effective acoustic representations from limited Taiwanese data alone.
Whisper-base, benefiting from large-scale multilingual pretraining, shows significant improvement and robust generalization across languages.
English-pretrained HuBERT-base offers moderate gains, indicating that cross-lingual transfer helps but is constrained by the phonological mismatch between English and Taiwanese Hokkien.
Compared with these strong baselines, CLiFT-ASR consistently reduces CER across all evaluation sets, achieving up to 26.2\% relative improvement on the development set and 24.88\% on the test set.
While Whisper-small slightly outperforms CLiFT-ASR on certain splits, it contains more than twice the number of parameters.
CLiFT-ASR therefore offers a parameter-efficient solution with substantial gains over competitive baselines, demonstrating the effectiveness of cross-lingual initialization combined with progressive two-stage fine-tuning.

\begin{figure*}[t]
\centering
\includegraphics[width=0.95\linewidth]{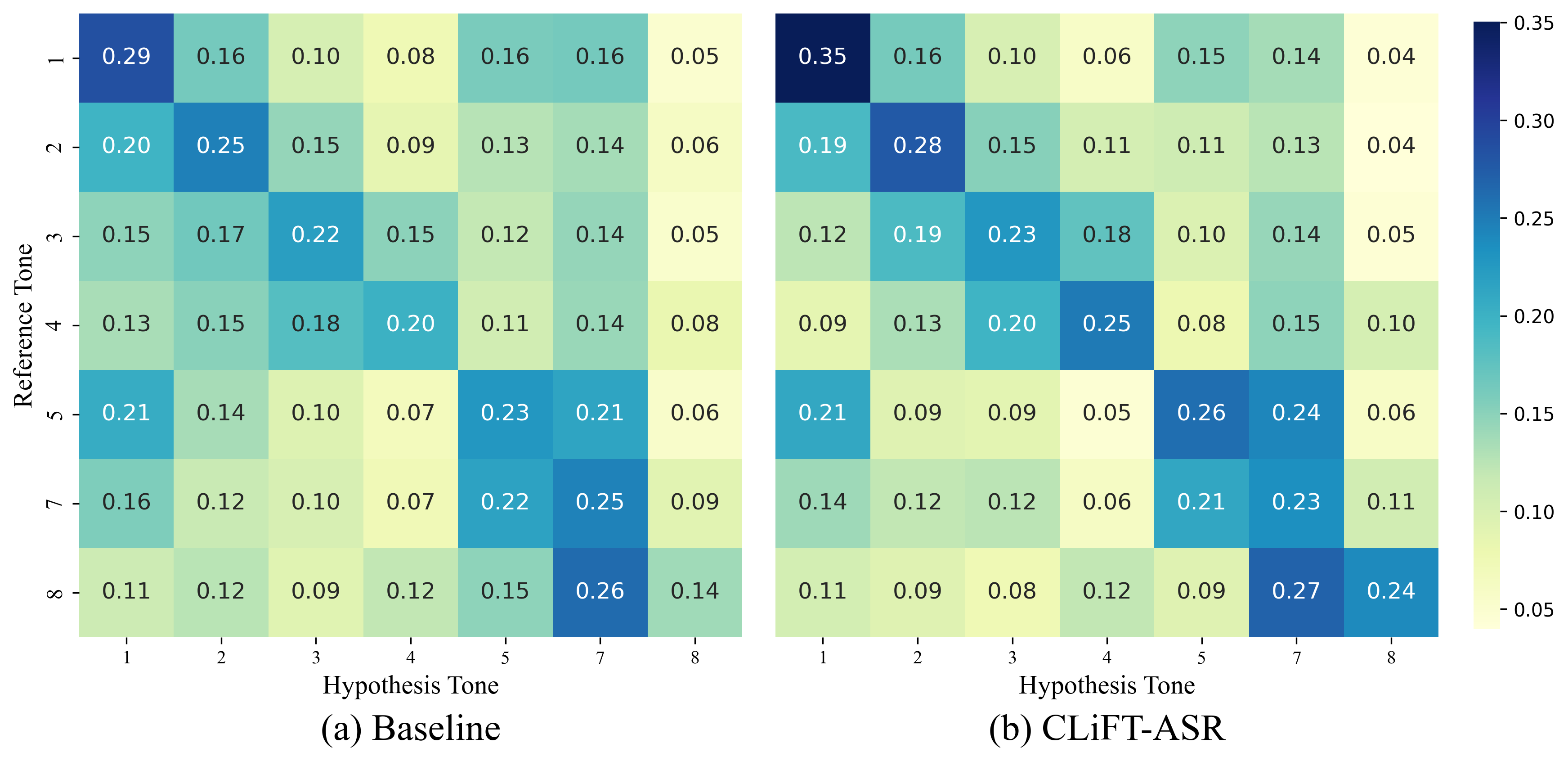}
\vspace{-5pt}
\caption{
Row-normalized substitution confusion matrices for Taiwanese Hokkien tone prediction, comparing the baseline and proposed models.
Tone labels are derived from Taibun surface forms without applying sandhi rules.
}
\label{fig:confusion}
\vspace{-10pt}
\end{figure*}

\subsection{Analysis of Fine-tuning Strategies}

Table~\ref{tab:finetune} presents the effects of different fine-tuning strategies and parameter freezing configurations on CLiFT-ASR performance.
Compared with direct end-to-end fine-tuning, the proposed two-stage strategy, which first adapts the model on phonetic (romanized) transcriptions and then refines it with Han character targets, consistently improves recognition accuracy across all evaluation sets.

Analyzing parameter freezing during the first stage highlights the contribution of each module.
Freezing the audio encoder or joint network restricts the ability of the model to adapt to target phonetics and orthography, leading to notable performance degradation, whereas freezing the prediction network has a milder effect.
The lowest CER is achieved when all components are trainable, indicating that full model adaptation within the two-stage fine-tuning strategy enables effective integration of acoustic and linguistic knowledge.
These results demonstrate that CLiFT-ASR with a carefully designed multi-stage fine-tuning strategy outperforms direct adaptation and provides a robust solution for low-resource mixed-orthography ASR scenarios.

\subsection{Investigation of Tone Confusions}

Figure~\ref{fig:confusion} depicts the substitution confusion matrix for tone prediction in Taiwanese Hokkien. 
The diagonal dominance indicates that most tones are correctly classified, yet tones 5, 7, and 8 exhibit frequent mutual misclassifications. 
These errors are likely attributed to tone sandhi phenomena, overlapping pitch contours, and speaker-dependent prosodic variations, which complicate accurate tone modeling in ASR.
To conduct this analysis, we employed the Taibun tool\footnote{\url{https://github.com/andreihar/taibun/}} to convert Taiwanese Han character outputs into Romanized forms with numerical tone labels. 
By aligning reference and predicted tone sequences, we constructed row-normalized substitution matrices to quantify tone-level confusions.

In the Zipformer baseline, tones 1, 5, and 7 emerge as the most error-prone categories. 
Tone 1 is correctly recognized only 29\% of the time, with 16\% of its instances misclassified as tone 2. 
Tone 5 is frequently misclassified as tone 1 (21\%), while tones 7 and 8 show substantial cross-confusions, indicating the limited ability of the baseline model to discriminate between acoustically similar tones.
In contrast, the proposed CLiFT-ASR system demonstrates clear improvements across most tonal categories. 
Tone 1 accuracy increases from 29\% to 35\%, while tone 4 recognition improves from 20\% to 25\%. 
The overall misclassification rate decreases, particularly for tones 5 and 7, where cross-tone errors are substantially reduced. 
These results highlight the enhanced discriminative capability of the proposed framework.
In summary, the tone confusion analysis confirms that CLiFT-ASR effectively reduces inter-tone errors, especially among acoustically similar tone pairs. 
This improvement can be attributed to the proposed feature design and training strategy, which together provide more robust tonal modeling for Taiwanese Hokkien ASR.

\section{Conclusion}

This study presents CLiFT-ASR, a cross-lingual fine-tuning framework designed for low-resource Taiwanese Hokkien ASR.
By initializing the audio encoder with Mandarin speech representations and applying an effective two-stage fine-tuning strategy, CLiFT-ASR achieves the best overall performance.
The first stage leverages Taiwanese romanization to capture detailed phonetic information, and the second stage adapts to Han character transcriptions to integrate orthographic and syntactic knowledge.
This progressive strategy highlights the advantage of aligning acoustic and linguistic representations in stages rather than directly training with limited annotated data.
An analysis of tone recognition shows that while general tone recognition is accurate, tones 5, 7, and 8 remain difficult due to tone sandhi, overlapping acoustic patterns, and speaker-specific prosodic variation, all of which complicate precise tone modeling.

\section{Future Work}

Several directions can be explored to extend the proposed CLiFT-ASR.
One promising avenue is targeted data augmentation that balances underrepresented tones.
Another is explicit modeling of tone sandhi, which may further reduce tonal confusion.
The integration of larger and more diverse pretraining corpora is expected to improve robustness, particularly for conversational speech.
Future research may also apply advanced sequence modeling or structured prediction techniques to capture tonal dependencies more effectively.
Finally, evaluating multilingual models such as Whisper could provide additional gains through large-scale pre-training and enhanced contextual modeling.

\section{Limitations}

Although CLiFT-ASR achieves competitive improvements, the current design relies on a stateless RNN-Transducer framework.
The stateless prediction network constrains the ability to model long-range dependencies, which may reduce accuracy in recognizing tonal patterns and complex tone sandhi.
Compared with recent large-scale pretrained models, the architecture also has limited capacity to exploit fully contextualized acoustic representations.
These limitations suggest that adopting more expressive architectures with stronger context modeling could further advance Taiwanese Hokkien ASR.

% \newpage

\bibliography{references.bib}
\bibliographystyle{acl_natbib}

\end{document}